\documentclass[letterpaper]{article} 
\usepackage{aaai2027}  
\nocopyright
\usepackage[hyphens]{url}  
\usepackage{graphicx} 
\usepackage{natbib}  
\usepackage{caption} 
\usepackage{booktabs}
\usepackage{multirow}

\title{FidLIF: Parallel Training of High-Fidelity Spiking Neurons with Exact BPTT}

\author{
    Kaixiang He\textsuperscript{\rm 1},
    Yujie Wu\textsuperscript{\rm 1}\corresponding
}

\affiliations{
    \textsuperscript{\rm 1}Department of Computing, The Hong Kong Polytechnic University\\
    kaixiang.he@connect.polyu.hk,
    yu-jie.wu@polyu.edu.hk
}

\usepackage{xcolor}
\newcommand{\rev}[1]{#1}
\newcommand{\sout}[1]{}

\begin{document}

\maketitle

\begin{abstract}

Although spiking neural networks (SNNs) show promise in energy-efficient computing through event-driven computation and stateful temporal processing, their recurrent membrane dynamics render their training inherently sequential and inefficient for long sequences.
Existing methods achieve full-sequence parallelism in SNNs through two main strategies: redesigning or approximating spiking dynamics to avoid reset-coupled temporal dependencies, or relying on iterative relaxed fixed-point solvers that suffer from excessive memory costs and iteration-dependent backward overhead. 
To address these limitations, we present FidLIF, a novel framework that reconciles the incremental update and hard-reset behavior of standard spiking LIF neurons with parallel forward propagation and BPTT-equivalent backward propagation. 
FidLIF iteratively recovers the full-sequence membrane-potential trajectory in parallel. By exploiting the lower bidiagonal structure induced by the local recurrence of LIF neurons, FidLIF formulates each
forward trajectory correction and backward adjoint propagation as associative scans with an $O(\log T)$ temporal depth.
This formulation also supports input-dependent neuronal parameters and allows the same neuron to perform full-sequence parallel 
training and incremental step-by-step inference at deployment.
In our numerical convergence evaluation, FidLIF achieves a mean absolute error below $10^{-7}$ relative to the matched serial relaxed LIF dynamics, while yielding gradients numerically equivalent to exact BPTT.
Our Triton implementation further delivers favorable long-sequence latency and memory scaling, 
while achieving competitive performance on four sequential classification benchmarks.
By reconciling parallel training with step-by-step execution, FidLIF enables high-fidelity spiking dynamics to benefit from modern parallel hardware without compromising incremental inference at deployment.
\end{abstract}

\section{Introduction}
\newcommand{\yes}{\ensuremath{\surd}}
\newcommand{\no}{\ensuremath{\times}}

Spiking neural networks (SNNs)  integrate stateful neuronal dynamics with sparse, event-driven communication \cite{maass1997networks,roy2019towards}, enabling highly energy-efficient computation on neuromorphic hardware.  With recent advances in surrogate-gradient learning,  backpropagation through time (BPTT) \cite{werbos1990backpropagation} can be effectively applied to SNN training. This has allowed SNNs to achieve competitive performance across a growing range of static or short-sequence tasks, such as image classification and object tracking \cite{neftci2019surrogate,wu2018spatio}.

Despite the great potential, training SNNs for long-sequence tasks remains an open challenge due to linearly increasing computational and memory costs.  Take the standard leaky integrate-and-fire (LIF) spiking neurons as an example. Although the subthreshold dynamics of LIF neurons are linear and theoretically parallelizable, threshold-based spike generation introduces resets and alters the subsequent evolution. Since we cannot obtain the spikes and resets in advance, forward propagation has to proceed sequentially over time, resulting in training costs that grow with the number of time steps.

To improve SNN capabilities for long-sequence learning, several parallel training approaches have been explored in recent years. For instance, some methods, such as Parallel Spiking Neurons (PSN) \cite{PSN}, achieve parallelization by completely removing the reset mechanism to decouple temporal dependencies. Alternatively, Forward Parallel Training (FPT) \cite{FPT} treats the soft-reset LIF neuron as a trajectory-solving problem, evaluating all time steps in parallel via iterations. However, these previous works face significant trade-offs: modifying the core spiking dynamics sacrifices the temporal expressivity and hard-reset nature of standard SNNs, while existing iterative solvers require backpropagating through all solver iterations, incurring excessive memory overhead. Moreover, most of these methods sacrifice incremental step-by-step inference capabilities while achieving parallelization, which is disadvantageous for deployment. Therefore, achieving efficient parallel training without compromising the fidelity of recurrent LIF dynamics or deployment inference remains largely unsolved.

\begin{figure*}[t]
\centering
\includegraphics[width=1.0\textwidth]{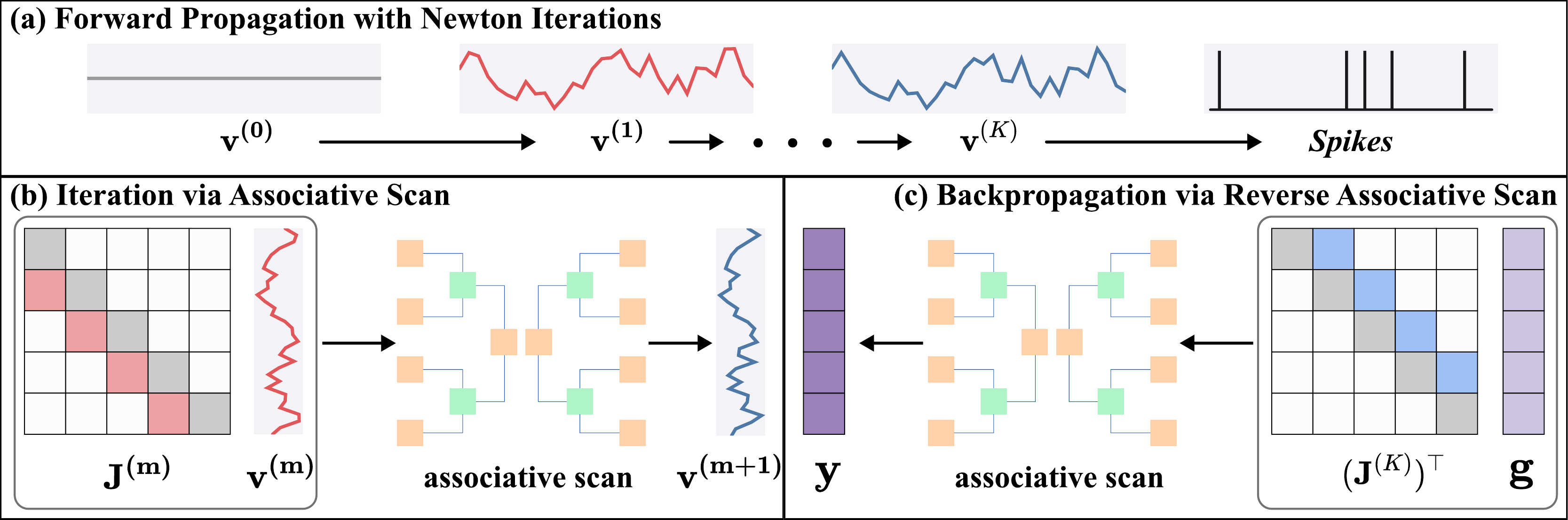}
\caption{Overall workflow of FidLIF. (a) In forward propagation, FidLIF uses Newton's method to recover the membrane potential $\mathbf{v}$ across all time steps. (b) Illustration of one forward iteration. Given the result of $m$-th iteration $\mathbf{v}^{(m)}$ and its corresponding Jacobian $\mathbf{J}^{(m)}$, the result of $(m+1)$-th iteration is obtained by a parallel scan. (c) In backpropagation, only one scan is needed. The transposed Jacobian matrix at the final trajectory is used to compute the adjoint vector $\mathbf{y}$ from upstream gradient $\mathbf{g}$.}
\label{overview}
\end{figure*}
To address these limitations, we propose FidLIF, a framework for reconciling parallel forward trajectory recovery and temporal credit assignment with high-fidelity recurrent hard-reset LIF neurons.
In forward propagation, FidLIF applies Newton iterations to a \rev{differentiable} trajectory equation, \rev{where the Heaviside function is relaxed by a sigmoid during the iterative solve.} At each iteration, linearizing the causal LIF dynamics yields associative temporal updates, allowing all time steps to be updated in parallel with a scan. \rev{After the trajectory is recovered, binary spikes are produced using the original Heaviside gate, preserving binary spike communication.} For the backward pass, we derive a reverse scan that is equivalent to BPTT when evaluated on the converged trajectory, avoiding backpropagation through iterations. To enhance modeling capacity without sacrificing incremental step-by-step execution, FidLIF causally generates input-dependent neuronal parameters from recent inputs. Consequently, FidLIF supports full-sequence parallel training and step-by-step incremental inference using the same recurrent neuron formulation.

Our contributions are as follows:
\begin{itemize}
 \item We introduce FidLIF, a novel framework that retains high-fidelity hard-reset dynamics of LIF neurons while enabling full-sequence parallel training and incremental step-by-step inference within the same neuron. FidLIF also
  supports input-dependent neuronal parameters for enhanced expressivity.
 \item We reformulate both forward and backward propagation as associative scans: membrane potential and spike trajectories are computed 
 through iterative parallel scans, 
 while exact BPTT temporal credit assignment is obtained through a single reverse scan, 
achieving $O(K\log T)$ and $O(\log T)$ temporal depth.
 \item We provide a Triton-based implementation for the proposed framework and comprehensively evaluate its performance across sequential image, event-vision, and speech classification tasks. Our results demonstrate that FidLIF delivers significant speedups and memory savings on long sequences,  while achieving competitive performance on versatile downstream tasks.
\end{itemize}

\section{Related Work}

\subsection{Parallel Methods for Spiking Neurons}

There are several parallel methods for spiking neurons. Some studies primarily focus on parallelizing subthreshold evolution by exploiting refractory periods or chunks between spikes \cite{alif,morrill2026bullet}. To parallelize the full sequence, another line of work redesigns or approximates spiking dynamics to avoid reset-related temporal dependencies. SPSN \cite{spsn} introduces stochastic spiking, separating the nonlinear spiking function and thus enabling parallel computing. PSN \cite{PSN} removes the reset of the spiking neuron and leverages a temporal matrix to capture temporal dependency. Its variants, Masked-PSN and sliding PSN, only maintain a window of a limited number of steps and thus capture only recent steps. PSU \cite{PSU} points out the importance of the reset mechanism and estimates the reset via a parallelizable reset estimation component. Building upon PSN, MF-PSN \cite{MFPSN} also removes the reset and further designs channel-wise temporal convolutions, using bit shifting to replace multiplication operations.
The Dynamic Decay Spiking Neuron \cite{DDSN} analyzes the function of reset and, based on this, redesigns the neuronal dynamics to achieve parallelization. 
PMSN \cite{pmsn} uses state space models (SSMs) to model multi-compartment spiking neurons and achieves parallelization via temporal-decoupled reset. SpikingSSMs \cite{spikingssm} uses a surrogate dynamic network to estimate the LIF dynamics for acceleration.

Although these methods are effective, most of them achieve parallelization by modifying or approximating the neuronal dynamics to eliminate temporal dependencies, which may compromise fidelity to standard LIF dynamics.

\subsection{Fixed-Point Iteration and Associative Scan}

Fixed-point iteration has been widely used in deep learning. These methods solve for the equilibrium states (fixed points) of the network and use implicit differentiation for training \cite{DEQ, ImplicitDL}. This idea is then adopted in SNNs with feedback, where the average firing rate serves as the equilibrium state \cite{ide,spide, MPIS}. 
Another approach treats the entire state trajectory of the recurrent neural network (RNN) as a fixed-point problem, thus enabling parallel evaluation over the sequence \cite{DEER,DeepPCR,ELK}. In SNNs, FPT \cite{FPT} is the first to apply this approach to spiking neuron parallelization. Unlike methods that take average firing rate as the equilibrium state, FPT frames the membrane potential trajectory across all time steps as the fixed-point and solves it through iteration, achieving parallelization of the soft-reset LIF dynamics.

However, FPT mainly considers soft-reset LIF. The forward propagation iteration is merely repeated substitution, and its backpropagation goes through the iterative process in the forward pass. Therefore, its forward iteration may converge slowly, and the memory consumption increases with the number of iterations, limiting its use in long sequences.

Associative scan is an algorithm that computes sequences of associative operations in logarithmic depth \cite{blelloch1990prefix}. It has been widely used to accelerate the linear RNN and SSMs \cite{DEER,s5,Mamba,dualprefix}. It has also been shown to be applicable for computing BPTT in RNNs \cite{bppsa}. In SNNs, it has been used for parallel computation of the subthreshold dynamics of continuous SNNs \cite{morrill2026bullet}. However, these studies do not use scans to solve discrete hard-reset LIF dynamics or their backpropagation.

\section{Methods}
We first introduce the LIF dynamics, then derive FidLIF’s parallel forward and backward passes, and finally establish finite-step convergence and BPTT equivalence.

\subsection{LIF Dynamics}

A widely used form of continuous LIF dynamics is \cite{standLIF}:

\begin{equation}
 \tau \frac{\mathrm{d} v(t)}{\mathrm{d}t}
 =
 -\left(v(t)-v_{\mathrm{rest}}\right)
 +
 R c(t)
\end{equation}

where $v(t)$ and $c(t)$ are the membrane potential and input current at time $t$, $v_{\mathrm{rest}}$ is the
resting potential, $\tau$ is the membrane time constant, and $R$ is the membrane resistance.

Using exponential discretization with a hard reset at the end of each time step, the discrete LIF dynamics are: 

\begin{equation}
\begin{array}{rcl}
s_{i}
& = &
H\left(v_{i}-v_{\mathrm{th}}\right), \\[2pt]
v'_{i}
& = &
\left(1-s_{i}\right)v_{i}
+
s_{i}v_{\mathrm{rest}}.
\end{array}
\label{eq:hard_reset}
\end{equation}

\begin{equation}
\begin{array}{rcl}
\rho
&=&
\displaystyle\exp\left(-\frac{\Delta t}{\tau}\right),\\
v_{i+1}
&=&
\displaystyle\rho
v'_i
+
(1-\rho)
\bigl(
v_{\mathrm{rest}}+Rc_{i+1}
\bigr) \\

&\stackrel{\triangle}{=}&

\displaystyle\phi_{i+1}(v_i;c_{i+1},\theta),
\end{array}
\label{eq:discrete_lif}
\end{equation}

where $i$ is the discrete time index, $s$ is the
spike, $v_{\mathrm{th}}$ is the firing threshold and $H$ is the Heaviside function. The quantity
$\Delta t$ denotes the discretization step size, $v'$ is the
post-reset membrane potential, and $\rho$ is the
corresponding decay factor. For brevity, we define $\phi_{i+1}$ to denote one-step transition, and $\theta$ to represent all the neuronal parameters.

Equations (\ref{eq:hard_reset}) and (\ref{eq:discrete_lif}) define the LIF dynamics that our method aims to parallelize. We hereafter refer to Eqs. (\ref{eq:hard_reset}) and (\ref{eq:discrete_lif}) as the
\emph{Heaviside-gated hard-reset LIF}.

\subsection{Parallel Forward}
Let $z_i = (1-\rho)(v_{\mathrm{rest}} + R c_i)$. Stacking Eqs. (\ref{eq:hard_reset}) and (\ref{eq:discrete_lif}) over all time steps yields:

\begin{equation}
\mathbf{v}=\mathbf{P}( (1-\mathbf{s})\odot \mathbf{v}+v_{rest}\mathbf{s})+\mathbf{q},
\label{eq:vector_ours}
\end{equation}
in which $\mathbf{v}$ and $\mathbf{s}$ are column vectors created by stacking the membrane potentials $\{v_i\}$ and spikes $\{s_i\}$ for $T$ time steps from $i=1$ to $i=T$, and
\begin{equation}
\mathbf{P}
=
\left(
\begin{array}{@{}ccccc@{}}
0     & 0     & \cdots & 0     & 0\\
\rho  & 0     & \cdots & 0     & 0\\
0     & \rho  & \ddots & 0     & 0\\
\vdots& \ddots& \ddots & \ddots& \vdots\\
0     & \cdots& 0      & \rho  & 0
\end{array}
\right),
\mathbf{q}
=
\left(
\begin{array}{@{}c@{}}
\rho v'_0+z_1\\
z_2\\
\vdots\\
z_T
\end{array}
\right),
\label{eq:P_and_q}
\end{equation}
$v'_0$ is the given initial value. We move the terms of the equation to the same side and get $\mathbf{G}(\mathbf{v})$:
\begin{equation}
\mathbf{G}(\mathbf{v})
=
\mathbf{v}
-
\mathbf{P}
\left(
\left(
\mathbf{1}-\mathbf{s}
\right)
\odot
\mathbf{v}
+
v_{\mathrm{rest}}\mathbf{s}
\right)
-
\mathbf{q}
=
\mathbf{0}.
\label{eq:forward_residual}
\end{equation}

 The membrane-potential trajectory $\mathbf{v}$ is regarded as the root of Eq. (\ref{eq:forward_residual}). We use Newton's iteration method to find the root.
At the $(m+1)$-th iteration, the Newton update $\Delta\mathbf{v}^{(m+1)}$ is obtained from
\begin{equation}
\begin{array}{rcl}
\mathbf{J_G}\bigl(\mathbf{v}^{(m)}\bigr)\,\Delta\mathbf{v}^{(m+1)}
&=& -\mathbf{G}\bigl(\mathbf{v}^{(m)}\bigr),\\
\mathbf{v}^{(m+1)} &=& \mathbf{v}^{(m)} + \Delta\mathbf{v}^{(m+1)},
\end{array}
\label{eq:newton_update}
\end{equation}
where $\mathbf{J}_{G}$ denotes the Jacobian of $\mathbf{G}$ with
respect to the entire membrane-potential trajectory $\mathbf{v}$.

The Heaviside function is non-differentiable. Following FPT, we replace it with a sigmoid during the iterations \cite{FPT}:
\[
s_i = \sigma\bigl(\alpha (v_i - v_{\mathrm{th}})\bigr),
\]
where $\sigma(\cdot)$ is the sigmoid function and $\alpha$ controls the steepness of the sigmoid. This substitution not only makes the equation differentiable but also stabilizes the iterative process. We refer to the LIF with sigmoid in place of Heaviside as \emph{sigmoid-relaxed LIF}.

Since $\mathbf{P}$ is strictly subdiagonal, its $i$-th row selects
only the post-reset state at the previous time step. Therefore, Eq.~(\ref{eq:forward_residual}) can be written component-wise as
\begin{equation}
G_i(\mathbf{v})
=
v_i-\phi_i(v_{i-1};c_i, \theta),
\quad i=1,2, \ldots,T.
\label{eq:component_residual}
\end{equation}
Therefore, $G_i$ depends only on the current state $v_i$ and the
immediately preceding state $v_{i-1}$. 
Consequently, the non-zero entries of the Jacobian $\mathbf{J_G}$ exist only on the main diagonal and the lower subdiagonal:
\begin{equation}
\left.
\frac{\partial G_i}{\partial v_j}
\right|_{\mathbf{v}=\mathbf{v}^{(m)}}
=
\left\{
\begin{array}{ll}
1,
& j=i,
\\
-a_j^{(m)},
& j=i-1,
\\
0,
& \mathrm{otherwise},
\end{array}
\right.
\label{eq:jacobian_elements}
\end{equation}

\begin{equation}
a_j^{(m)}
=\left.
\frac{\partial \phi_{j+1}(v_j;c_{j+1},\theta)}{\partial v_j}
\right|_{v_j=v_j^{(m)}}.
\label{eq:local_newton_coefficient}
\end{equation}

Let $\mathbf{A}^{(m)}$ denote the strictly subdiagonal part, i.e.,
\begin{equation}
A_{ij}^{(m)}
=
\left\{
\begin{array}{ll}
a_j^{(m)}, & i=j+1,\\
0, & \mathrm{otherwise}.
\end{array}
\right.
\label{eq:newton_A}
\end{equation}
Accordingly, the Jacobian can be written as
\begin{equation}
\mathbf{J}_{G}
\left(
\mathbf{v}^{(m)}
\right)
=
\mathbf{I}
-
\mathbf{A}^{(m)}.
\label{eq:jacobian_compact}
\end{equation}
Substituting Eq.~(\ref{eq:jacobian_compact}) into the Newton method Eq.~(\ref{eq:newton_update})
and rearranging, we have
\begin{equation}
\mathbf{v}^{(m+1)}
=
\mathbf{A}^{(m)}
\mathbf{v}^{(m+1)}
+
\mathbf{b}^{(m)},
\label{eq:newton_affine_system}
\end{equation}
\begin{equation}
\mathbf{b}^{(m)}
=
\left(
\mathbf{I}
-
\mathbf{A}^{(m)}
\right)
\mathbf{v}^{(m)}
-
\mathbf{G}
\left(
\mathbf{v}^{(m)}
\right).
\label{eq:newton_b}
\end{equation}
Because $\mathbf{A}^{(m)}$ is strictly subdiagonal, expanding Eq. (\ref{eq:newton_affine_system}) component-wise yields:
\begin{equation}
\begin{array}{rcl}
v_1^{(m+1)}
&=&
b_1^{(m)},
\\
v_i^{(m+1)}
&=&
a_{i-1}^{(m)}
v_{i-1}^{(m+1)}
+
b_i^{(m)},
\qquad
i=2,\ldots,T.
\end{array}
\label{eq:newton_affine_recurrence}
\end{equation}

\sout{Eq. (\ref{eq:newton_affine_recurrence}) is evaluated using an associative scan with $O(\log T)$ depth. The scan operator and its proof of associativity are provided in the supplementary material.
}
\rev{Eq. (\ref{eq:newton_affine_recurrence}) is an affine recurrence. By substituting two consecutive updates, we obtain\[v_{i+1}^{(m+1)}
=
a_i^{(m)}a_{i-1}^{(m)}v_{i-1}^{(m+1)}
+
a_i^{(m)}b_i^{(m)}
+
b_{i+1}^{(m)}.\]
Thus, composing two affine updates amounts to combining their
coefficient pairs according to
\[
(x_2,y_2)\circ(x_1,y_1)
:=
(x_2x_1,\;x_2y_1+y_2).
\]
For Eq. (\ref{eq:newton_affine_recurrence}), each temporal update corresponds to
$(a_{i-1}^{(m)}, b_i^{(m)})$. Since this composition is associative,
all prefix updates can be evaluated by an associative scan with
$O(\log T)$ temporal depth.}

After $K$ iterations, the recovered membrane potential trajectory
$\mathbf{v}^{(K)}$ yields the spike sequence via the Heaviside gate
$\mathbf{s} = H(\mathbf{v}^{(K)} - v_{\mathrm{th}})$
as in Eq.~(\ref{eq:hard_reset}). \sout{The final results are therefore binary spikes.} \rev{Thus, although a smooth relaxation is used during trajectory
recovery, the final neuron outputs remain binary spikes, preserving the spike representation of standard LIF neurons.}

\subsection{Parallel Backward Credit Assignment}
Instead of backpropagating through the iterations,
we perform implicit differentiation on the equation:
\begin{equation}
\mathbf{G}
\left(
\mathbf{v}^{\star};
\mathbf{c},
{\theta}
\right)
=
\mathbf{0}.
\label{eq:implicit_fixed_point}
\end{equation}
The superscript
$\star$ indicates the converged fixed-point trajectory from the
forward pass. $\mathbf{c}$ denotes the input currents and $\mathbf{\theta}$ represents all trainable neuronal parameters.
Differentiating the equation with respect to
any variable $\mathbf{x}\in\{\mathbf{c},\theta\}$ gives
\begin{equation}
\frac{\partial\mathbf{v}^{\star}}{\partial\mathbf{x}}
=
-
\mathbf{J}_{G}
\left(
\mathbf{v}^{\star}
\right)^{-1}
\frac{\partial\mathbf{G}}{\partial\mathbf{x}}.
\label{eq:implicit_diff}
\end{equation}
Let
$\mathbf{g}=\nabla_{\mathbf{v}^{\star}}\mathcal{L}$ denote the direct upstream gradient of $\mathbf{v^{\star}}$, where $\mathcal{L}$ is the loss function.
Assuming that $\mathcal{L}$ depends on $\mathbf{x}$ through
$\mathbf{v}^{\star}$, the chain rule and Eq. (\ref{eq:implicit_diff}) give
\begin{equation}
\nabla_{\mathbf{x}}\mathcal{L}
=
\displaystyle
-
\left(
\frac{\partial\mathbf{G}}
{\partial\mathbf{x}}
\right)^{\mathsf{T}}
\mathbf{J}_{G}
\left(
\mathbf{v}^{\star}
\right)^{-\mathsf{T}}
\mathbf{g}.
\label{eq:implicit_loss_gradient}
\end{equation}
We therefore define the \emph{adjoint vector}
$\mathbf{y}$ by
\begin{equation}
\mathbf{J}_{G}
\left(
\mathbf{v}^{\star}
\right)^{\mathsf{T}}
\mathbf{y}
=
\mathbf{g}.
\label{eq:adjoint_system}
\end{equation}
The upstream gradient $\mathbf{g}$ is given by the subsequent
layers, while $\partial\mathbf{G}/\partial\mathbf{x}$ can be
computed across all time steps at the fixed trajectory
$\mathbf{v}^{\star}$ . Neither requires sequential propagation over time.
Therefore, the only temporally coupled quantity to be solved is the
adjoint vector $\mathbf{y}$ in Eq. (\ref{eq:adjoint_system}).

From the forward derivation,
$\mathbf{J}_{G}(\mathbf{v}^{\star})
=\mathbf{I}-\mathbf{A}^{\star}$,
where $\mathbf{A}^{\star}$ is strictly lower subdiagonal.
Therefore, $\mathbf{J}_{G}(\mathbf{v}^{\star})^{\mathsf{T}}$ is upper
bidiagonal. Similar to Eq. (\ref{eq:newton_affine_recurrence}), we can write Eq. (\ref{eq:adjoint_system}) as
\begin{equation}
\begin{array}{rcl}
y_T
&=&
g_T,
\\
y_i
&=&
a_i^{\star}y_{i+1}+g_i,
\qquad
i=T-1,\ldots,1.
\end{array}
\label{eq:adjoint_affine_recurrence}
\end{equation}
This is the same affine transformation as in the forward iterations, but in
reverse order. It can therefore be computed by a reverse
associative parallel scan.

Notably, the adjoint scan requires only the trajectory
$\mathbf{v}^{\star}$ and the per-step coefficients
$\{a_i^{\star}\}_{i=1}^{T-1}$; it does not depend on the forward iterates $\mathbf{v}^{(m)}$ nor on the forward solver's computation process. Therefore, the forward iteration can be performed with gradient tracking disabled without affecting backpropagation, decoupling memory consumption from the number of iterations $K$.

\subsection{Input-Dependent Neuron Parameters}
To enhance the expressive capacity of the neuron, we make the
time constant $\tau_i$ and resistance $R_i$ adaptive to the recent input
context, which is inspired by recent studies in sequence models \cite{Mamba,rwkv}. Let
$\mathcal C_i=(c_{i-n+1},\ldots,c_i)$ denote a causal window containing
the most recent $n$ inputs. A small causal parameter generator
$g_{\psi}$ gives
\begin{equation}
    (\Delta\tau_i,\Delta R_i)=g_{\psi}(\mathcal C_i).
\end{equation}
To constrain the range of parameters and improve training stability, we define the final dynamics parameters as follows:
\begin{equation}
\begin{array}{rcl}
\tau_i
&=&
\tau_{\min}+(\tau_{\max}-\tau_{\min})
\sigma(\tau_{\mathrm{base}}+\Delta\tau_i),\\
R_i
&=&
R_{\min}+(R_{\max}-R_{\min})
\sigma(R_{\mathrm{base}}+\Delta R_i).
\end{array}
\end{equation}
The bounds $\tau_{\min},\tau_{\max},R_{\min},R_{\max}$ and the base values 
$\tau_{\mathrm{base}},R_{\mathrm{base}}$ are tunable hyperparameters; 
we always constrain $\tau_i>0$ so that $\rho_i = \exp(-\Delta t/\tau_i) < 1$, 
preserving the membrane potential decay characteristic of the LIF model.

\sout{All $\{\tau_i,R_i\}_{i=1}^{T}$ are generated before the iterative solve
and remain fixed across solver iterations. Since they depend only on the
input rather than the solver iterates, neither the iterative method nor backpropagation mentioned above is affected. Gradients with respect to $g_{\psi}$ can be obtained through standard automatic
differentiation.}

\rev{
All $\{\tau_i, R_i\}_{i=1}^{T}$ are generated from causal input contexts before the iterative solve, so no future inputs are required and step-by-step inference is preserved. These parameters remain fixed throughout the $K$ Newton iterations; therefore, they do not introduce additional temporal dependencies into the solver, and the same parallel scan formulation remains applicable. During backpropagation, gradients with respect to $\tau_i$ and $R_i$ are computed by the adjoint backpropagation derived above, and gradients are then propagated through $g_\psi$ using standard automatic differentiation.
}
\sout{
Training-time pseudocode and parameter-generator details are provided in the supplementary material.}
\begin{figure*}[t]
\centering
\includegraphics[width=1.0\textwidth]{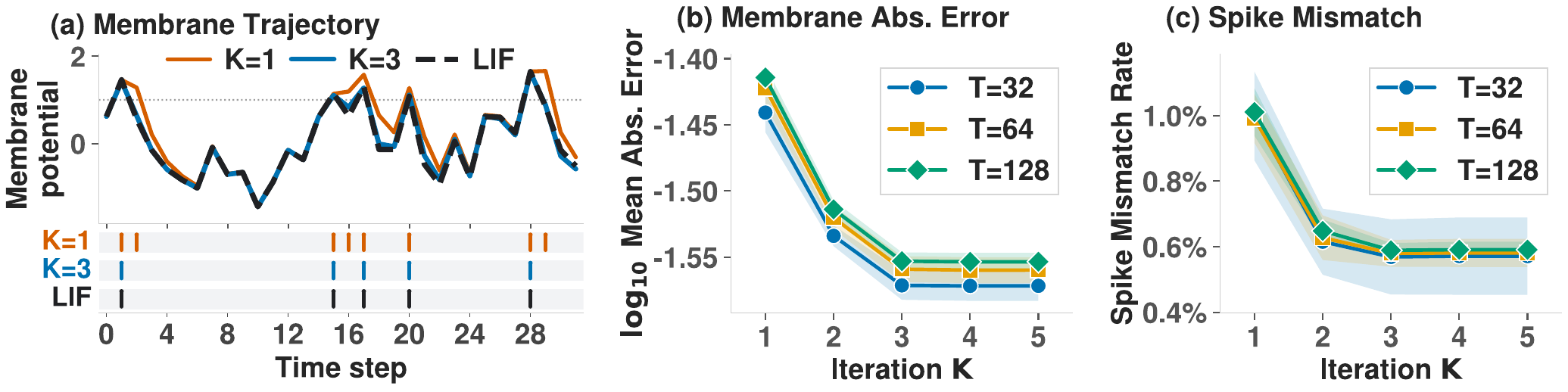}
\caption{Fidelity to Heaviside-gated hard-reset LIF dynamics.
(a) Representative membrane-potential (top) and spike trajectories (bottom) for $T=32$, comparing serial LIF with FidLIF using $K=1$ and $3$ iterations.
(b) Mean absolute error (MAE) of membrane potential.
(c) Spike mismatch rate relative to LIF. Curves in (b) and (c) show the mean over five independent trials, and shaded regions denote one standard deviation.}
\label{fig:FidLIF_fidelity}
\end{figure*}

\subsection{Theoretical Analysis}

Let
$v_i^\star=\phi_i(v_{i-1}^\star;c_i,\theta)$
denote the sequential solution. Using the definition of
$b_i^{(m)}$, Eq. (\ref{eq:newton_affine_recurrence}) can be written as
\begin{equation}
\begin{array}{rcl}
v_1^{(m+1)}
&=&
\displaystyle\phi_1(v_0;c_1,\theta),\\
v_i^{(m+1)}
&=&
\displaystyle\phi_i(v_{i-1}^{(m)};c_i,\theta)
+
a_{i-1}^{(m)}
\bigl(
v_{i-1}^{(m+1)}-v_{i-1}^{(m)}
\bigr), \\
\quad && i=2,\ldots,T.
\end{array}
\label{eq:anchored_scalar_update}
\end{equation}
Equation~(\ref{eq:anchored_scalar_update}) explains why Newton's method can converge faster than the repeated substitution
used in FPT. FPT computes every state in the $(m+1)$-th
iteration solely from the previous iteration $\mathbf v^{(m)}$.
Therefore, a newly updated state cannot affect time steps after it until the next iteration.

Newton's method instead estimates the latest changes in the current iteration using first-order derivatives. In Equation (\ref{eq:anchored_scalar_update}), the first term computes the dynamics using the previous iteration, while the second term estimates the change caused by the latest update
$v_{i-1}^{(m+1)}-v_{i-1}^{(m)}$. If this estimation is accurate, one Newton iteration can approximate and propagate the effect of the latest updates through multiple time steps. 
The following theorem further guarantees exact recovery of the trajectory within at most $T$ iterations.

\textbf{Theorem 1 (Finite-step convergence).~} In exact arithmetic, Eq.~(\ref{eq:anchored_scalar_update}) recovers
the membrane potential trajectory within at most $T$ iterations for any
initialization, provided that all coefficients remain finite.
More precisely, after iteration $m$,
$v_i^{(m)}=v_i^\star$ for every $i\le m$.

\emph{Proof sketch.}
We prove by induction on $m$. 
Since $v_0$ is fixed, the first line of Eq.~(\ref{eq:anchored_scalar_update}) 
gives $v_1^{(1)}=\phi_1(v_0;c_1,\theta)=v_1^\star$.
For the inductive step, suppose $v_i^{(m)} = v_i^\star$ for all $i \leq m$.
At iteration $m+1$, for each $i=2,\ldots,m+1$, the forward recurrence propagates 
$v_{i-1}^{(m+1)} = v_{i-1}^{(m)}$ step by step; the correction term 
$a_{i-1}^{(m)}(v_{i-1}^{(m+1)}-v_{i-1}^{(m)})$ therefore stays zero, 
so $v_i^{(m+1)} = \phi_i(v_{i-1}^\star;c_i,\theta) = v_i^\star$.
Thus, every iteration extends the correct prefix by at least
one time step; the sequential trajectory is recovered within at most $T$ iterations, 
i.e., $\mathbf v^{(T)}=\mathbf v^\star$.

\sout{A complete proof of Theorem 1 is provided in the supplementary material.} Empirically, the number of iterations $K \ll T$ suffices.

As for backpropagation, we can demonstrate that, given identical forward trajectories and single-step derivatives, our backpropagation method is equivalent to BPTT.

\paragraph{Equivalence to BPTT.}
Let
$\delta_i=d\mathcal L/dv_i^\star$
denote the total state gradient computed by BPTT, while
$g_i=\partial\mathcal L/\partial v_i^\star$
denotes the gradient excluding the temporal path through future
states. Applying the chain rule to
$v_{i+1}^\star=\phi_{i+1}(v_i^\star)$ gives
\begin{equation}
\delta_T=g_T,
\qquad
\delta_i
=
g_i+a_i^\star\delta_{i+1},
\quad
i=T-1,\ldots,1.
\label{eq:bptt_recurrence}
\end{equation}
This is identical to the recurrence in Eq. (\ref{eq:adjoint_affine_recurrence}). Thus,
$y_i=\delta_i$ for every time step.
Moreover, since
$G_i=v_i-\phi_i(v_{i-1};\mathbf x)$, we have
$\partial G_i/\partial\mathbf x
=-\partial\phi_i/\partial\mathbf x$.
Substituting $\mathbf y=\delta$ into the implicit
gradient yields the same input and parameter gradients as BPTT.
Therefore, the proposed adjoint backward pass is BPTT-equivalent as long as the forward trajectory is identical and the per-step derivative definition is consistent.

\section{Experiments}

\begin{figure}[t]
\centering
\includegraphics[width=1.0\columnwidth]{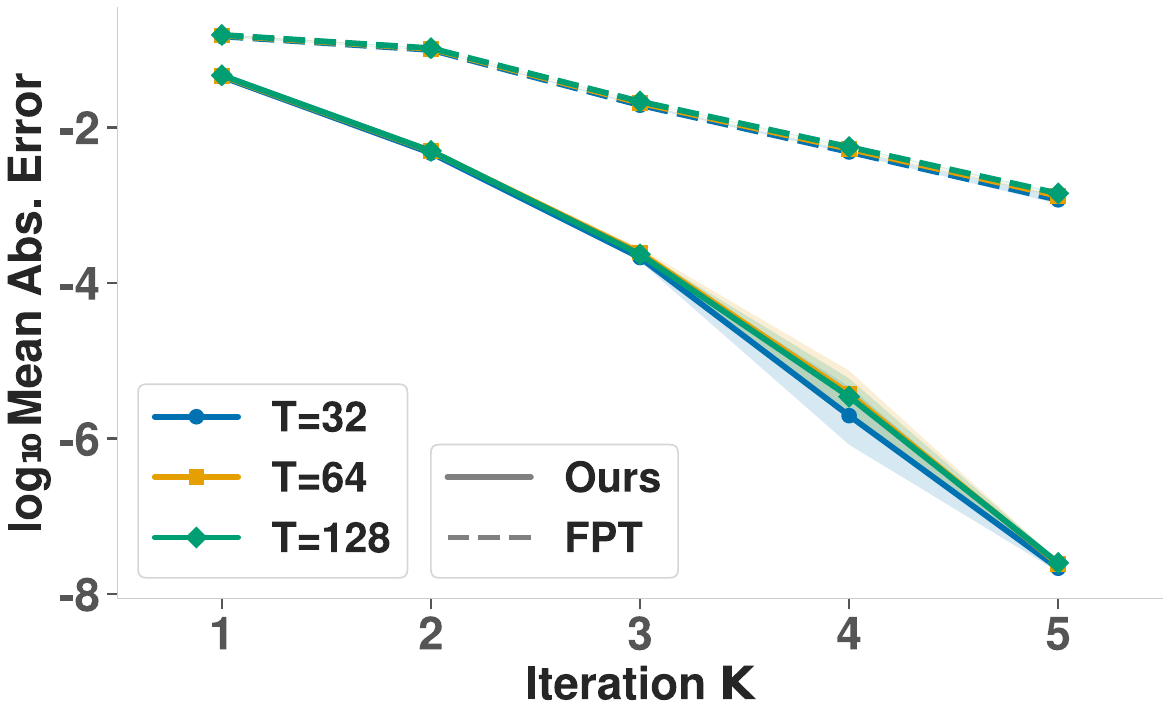}
\caption{Convergence to matched sigmoid-relaxed LIF dynamics.
Each method is compared against its own matched serial baseline.}
\label{fig:trajectory_comparison}
\end{figure}

\begin{figure*}[t]
\centering
\includegraphics[width=1.0\textwidth]{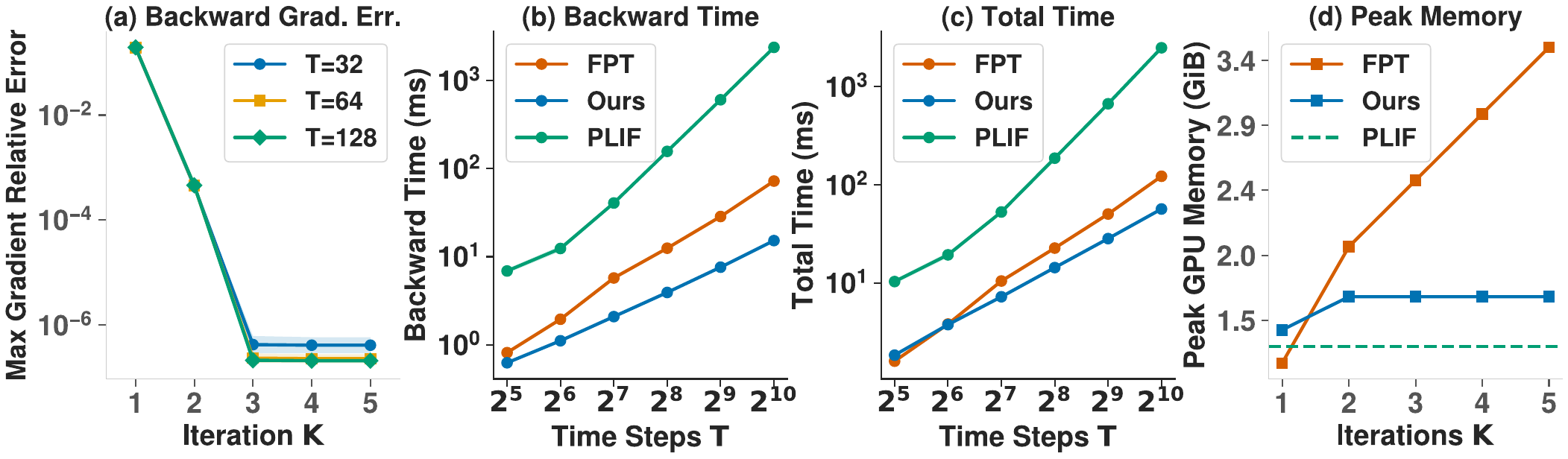}
\caption{Gradient recovery and computational efficiency.
(a) Gradient error vs.\ \rev{maximum} iteration budgets $K$ (mean$\pm$std, five seeds).
(b) Backward and (c) combined forward--backward latency at \rev{fixed iterations} $K=3$.
(d) Peak memory at $T =256$ vs.\ \rev{fixed iterations} $K$.}
\label{fig:performance}
\end{figure*}

\begin{table*}[t]
    \centering
    {
    \setlength{\tabcolsep}{5pt}
    \small
    \begin{tabular*}{\textwidth}{
    @{\extracolsep{\fill}}
    llcccl
    @{}}
        \toprule
        Dataset
        & Method
        & Arch.
        & \shortstack{Efficient Step-by-\\step inference}
        & \shortstack{Parallelization\\Strategy}
        & Accuracy (\%) \\
        \midrule

        \multirow{5}{*}{Seq. CIFAR10}

        & PSN / Sliding PSN
        & ConvNet 
        & \no{} / \yes$^{*}$
        & Reset Removal 
        & $88.45$ / $86.70$ \\
        
        & PSU / IPSU
        & ConvNet 
        & \yes{} / \no$^{\dagger}$
        & Reset Estimation 
        & $79.52$ / $87.28$ \\
        
        & MF-PSN
        & ConvNet 
        & \yes$^{*}$
        & Reset Removal
        & \underline{91.17} \\

        & FPT / FPT with $\mathbf{A}$
        & ConvNet 
        & \yes{} / \no$^{\dagger}$
        & Repeated Substitution
        & $81.54{\pm}0.34$ / $89.53{\pm}0.13$ \\
        
        & \textbf{FidLIF (Ours)}
        & ConvNet 
        & \yes
        & Newton's method + Scan
        & $90.99{\pm}0.25$ (\textbf{91.25}) \\

        \midrule

        \multirow{5}{*}{Seq. CIFAR100}

        & PSN / Sliding PSN
        & ConvNet 
        & \no{} / \yes$^{*}$
        & Reset Removal  
        & $62.21$ / $62.11$ \\
        
        & PSU / IPSU
        & ConvNet 
        & \yes{} / \no$^{\dagger}$
        & Reset Estimation
        & $51.65$ / $59.76$ \\
        
        & MF-PSN
        & ConvNet 
        & \yes
        & Reset Removal
        & \underline{66.21} \\

        & FPT / FPT with $\mathbf{A}$
        & ConvNet 
        & \yes{} / \no$^{\dagger}$
        & Repeated Substitution 
        & $55.19{\pm}0.37$ / $64.50{\pm}0.34$ \\

        & \textbf{FidLIF (Ours)}
        & ConvNet 
        & \yes
        & Newton's method + Scan
        & $66.42{\pm}0.17$ (\textbf{66.59}) \\

        \midrule

        \multirow{4}{*}{CIFAR10-DVS}
        & PSU / RPSU
        & VGG 
        & \yes{} / \no$^{\dagger}$
        & Reset Estimation
        & $81.40$ / $82.00$ \\
        
        & Sliding PSN
        & VGG 
        & \yes$^{*}$
        & Reset Removal
        & \textbf{85.90} \\
        
        & FPT with $\mathbf{A}$
        & VGG 
        & \no$^{\dagger}$
        & Repeated Substitution 
        & 85.50$\pm$0.22 (\underline{85.70}) \\
        
        & \textbf{FidLIF (Ours)}
        & VGG
        & \yes
        & Newton's method + Scan 
        & $84.13{\pm}0.65$ (84.8) \\

        \midrule

        \multirow{3}{*}{SHD}
        & PSU / IPSU / RPSU
        & FCNet 
        & \yes{} / \no$^{\dagger}$ / \no$^{\dagger}$
        & Reset Estimation
        & $90.64$ / $91.30$ / $92.49$ \\
        
        & MF-PSN
        & FCNet        
        & \yes$^{*}$
        & Reset Removal
        & \textbf{95.50$\pm$0.36} \\
        
        & \textbf{FidLIF (Ours)}
        & FCNet 
        & \yes
        & Newton's method + Scan
        & $94.38{\pm}0.31$ (\underline{94.68}) \\

        \bottomrule
    \end{tabular*}
    }
    \caption{
        Comparison with other full-sequence parallel SNNs.
        Error bars denote mean $\pm$ standard deviation, with the best
        run shown in parentheses. 
        $^{*}$: Fixed-length temporal window.
        $^{\dagger}$: Learnable temporal parameters break the recurrent structure.
    }
    \label{tab:task_performance}
\end{table*}

We evaluate FidLIF in four aspects: fidelity to Heaviside-gated hard-reset LIF dynamics, backward correctness and computational efficiency, task performance, and \sout{recurrent} \rev{incremental} inference. We further analyze the effects of the number of iterations and input-dependent neuronal parameters. \sout{Complete configurations and measurement methods are provided in the supplementary material.}

\subsection{Fidelity to LIF Dynamics}
We evaluate the fidelity of FidLIF on a single layer using Gaussian input currents $c \sim \mathcal{N}(0,1)$. Except for the representative trajectory, results are averaged over five trials with shaded regions denoting one standard deviation.

Fig. \ref{fig:FidLIF_fidelity} compares FidLIF directly with the Heaviside-gated hard-reset LIF dynamics. As the number of iterations increases, both the membrane potential error and the spike mismatch rate decrease. 
\rev{The residual error arises from two sources: the finite-iteration error in recovering the sigmoid-relaxed trajectory, and the discrepancy between the sigmoid-relaxed and original Heaviside-gated dynamics.} These results confirm that FidLIF faithfully recovers the dynamics of the Heaviside-gated hard-reset LIF within very few iterations.

\rev{To isolate the solver error from the discrepancy in the underlying dynamics,} Fig.~\ref{fig:trajectory_comparison} compares each method against its own matched serial sigmoid-relaxed baseline. FidLIF is compared against the sigmoid-relaxed hard-reset LIF while FPT is compared against the sigmoid-relaxed subtractive-reset LIF. The results show that FidLIF converges substantially faster, reducing the mean absolute error of the membrane potential below $10^{-7}$ with five iterations. This shows that, under the same iterations,\sout{FidLIF recovers the target dynamics more accurately.} \rev{the Newton-based FidLIF recovers the target relaxed dynamics more accurately than the repeated substitution method used in FPT.}

\subsubsection{Backward Correctness and Computational Efficiency}

Given an input current sequence, we use the gradients obtained via serial step-by-step forward propagation and automatic differentiation (i.e., BPTT) as a reference. Then, we run FidLIF on the same current sequence and compute backpropagation gradients using our adjoint scan after each iteration. Finally, we compare the gradient calculated after each iteration with the reference gradient. As shown in Fig. \ref{fig:performance}(a), the gradient error decreases rapidly with $K$, \rev{while larger values of $K$ may yield identical errors after convergence
due to tolerance-based early stopping.} At $K=3$, the gradient error is less than $10^{-6}$. \rev{Since the backward adjoint scan is BPTT-equivalent for an identical forward trajectory, the remaining gradient discrepancy at finite \(K\) mainly reflects incomplete forward-trajectory recovery.}

Finally, we test computational efficiency on a single layer with parameter generation disabled, using an RTX 4090 in FP32 (TF32 disabled). The implementation of PLIF comes from SpikingJelly \cite{plif}.
Figures~\ref{fig:performance}(b) and \ref{fig:performance}(c) report backward and combined forward--backward latency across $T$ at $K=3$. \rev{The latency advantage of FidLIF increases with $T$.} Figure~\ref{fig:performance}(d) shows peak memory across $K$ at $T=256$. While FPT's memory grows near-linearly, FidLIF maintains a nearly constant memory usage because it does not need per-iteration computation graphs.
\begin{table}[t]
    \centering
    {
    \small

    \setlength{\tabcolsep}{8pt}
        \begin{tabular*}{\columnwidth}{
    @{\extracolsep{\fill}}
    lcc
    @{}}
        \toprule
        Dataset
        & Parallel Acc. (\%)
        & Serial Acc. (\%) \\
        \midrule
        Seq. CIFAR10
        & $90.99\pm0.25$
        & $90.69\pm0.19$ \\

        Seq. CIFAR100
        & $66.42\pm0.17$
        & $65.98\pm0.24$ \\

        CIFAR10-DVS
        & $84.13 \pm 0.65$
        & $83.50 \pm 0.96$ \\
        
        SHD
        & $94.38 \pm 0.31$
        & $94.30 \pm 0.39$ \\

        \bottomrule
    \end{tabular*}
    }
    \caption{
        Accuracy comparison between parallel full-sequence execution
        and serial step-by-step execution.
}
    \label{tab:parallel_serial_consistency}
\end{table}
\begin{table}[t]
    \centering
    {
    \small
    \setlength{\tabcolsep}{6pt}
        \begin{tabular*}{\columnwidth}{
    @{\extracolsep{\fill}}
    rcc
    @{}}
        \toprule
        ~~$K$
        & Seq. CIFAR10 (\%)
        & Seq. CIFAR100 (\%) \\
        \midrule
        3
        & $90.99\pm0.25$
        & $66.42\pm0.17$ \\

        5
        & $90.79\pm0.20$
        & $66.57\pm0.42$ \\

        \bottomrule
    \end{tabular*}
    }
    \caption{
        Effect of the \rev{maximum} iterations $K$.
    }
    \label{tab:iteration_ablation}
\end{table}
\begin{table}[t]
    \centering
    {
    \small
    \setlength{\tabcolsep}{5pt}
    \begin{tabular*}{\columnwidth}{
    @{\extracolsep{\fill}}
    lcc
    @{}}
        \toprule
        Method
        & Seq. CIFAR10 (\%)
        & Seq. CIFAR100 (\%) \\
        \midrule
        FidLIF
        & $\mathbf{91.25}$
        & $\mathbf{66.59}$ \\
        w/o input-dependent
        & $80.66$
        & $54.52$ \\
        \bottomrule
    \end{tabular*}
    }
    \caption{
        Ablation of input-dependent neuronal dynamics.
    }
    \label{tab:input_conditioned_ablation}
\end{table}
\subsection{Task Performance}

We evaluate FidLIF on sequential CIFAR10/100, CIFAR10-DVS \cite{cifar10dvs}, and SHD \cite{shd}\sout{. All of them are widely used benchmarks in SNNs.}, \rev{ covering sequential images, event-based vision, and long-horizon spiking speech.} Table \ref{tab:task_performance} compares task accuracies and the features of different methods. The task accuracies for other methods are taken from \cite{PSN,MFPSN,PSU,FPT}, respectively. The ConvNet used for sequential CIFAR10/100 tasks and the VGG architecture used for CIFAR10-DVS task both use a modified version implemented by PSN \cite{PSN}. For SHD, we use the FCNet with learnable delays \cite{snndelay}, which is the same approach as MF-PSN \cite{MFPSN}.
All FidLIF task-performance experiments in Table \ref{tab:task_performance} \sout{use $K = 3$} \rev{set the maximum number of forward iterations to $K=3$} and are repeated three times, reported as mean $\pm$ std.

Most parallel neurons in Table~\ref{tab:task_performance} break the serial form, preventing incremental step-by-step inference. Methods using temporal matrices cannot be reduced to a recurrence because their entries are learned independently, breaking the structure of recurrence. Sliding PSN and MF-PSN restore serial ability by restricting temporal dependence to a finite window. Although original FPT and PSU retain a serial form, they do so by keeping their dynamic parameters fixed. In contrast, FidLIF supports full-sequence parallel training and step-by-step incremental inference. Moreover, the learned neuronal dynamics parameters in other methods are usually unconstrained, while FidLIF enforces $0<\rho_t<1$, preserving the leaky dynamics of LIF neurons. 

FidLIF is highly competitive. On sequential CIFAR10, FidLIF's best individual run reaches $91.25\%$. On sequential CIFAR100 FidLIF reaches $66.42\%$. On CIFAR10-DVS it attains $84.13\%$, and on SHD, which has 80–140 time steps, FidLIF reaches $94.38\%$, showing that \rev{a maximum of} three iterations remains effective even on sequences of such length.

\subsection{Incremental Inference and Ablation Studies}
We conduct three studies to examine different aspects of FidLIF: parallel-to-serial consistency, the sensitivity to the number of iterations $K$, and the contribution of input-dependent neuronal parameters.

\textbf{Parallel-to-Serial Consistency.}
A central advantage of FidLIF is that the same trained neuron supports both full-sequence parallel and step-by-step incremental form. We evaluate each trained model in both execution forms without retraining or fine-tuning. As shown in Table \ref{tab:parallel_serial_consistency}, the serial accuracy closely matches the parallel accuracy. \sout{The source of the gaps is mainly the error between the result of a finite number of iterations and the true fixed point.} \rev{The remaining gaps mainly reflect the discrepancy between the parallelly recovered dynamics and exact step-by-step serial dynamics. In multi-layer networks, such neuron-level discrepancies can accumulate across layers, while task-dependent activation distributions also differ from the Gaussian currents used in the single-layer fidelity experiment.} Overall, \rev{a maximum of three iterations is sufficient}\sout{three iterations are sufficient} for complex networks on real benchmarks, enabling models trained in parallel forward passes to be deployed for serial inference, which is essential for real-time edge applications.

\textbf{Effect of the \rev{Maximum} Number of Iterations.}
We also examine whether training benefits from more iterations. Table~\ref{tab:iteration_ablation} compares \rev{maximum iterations} $K=3$ and $5$ on the two sequential CIFAR benchmarks. Accuracy differs by less than $0.3\%$, indicating that the trajectories recovered by \rev{a maximum iteration budget of} $K=3$ are already enough for training. \sout{Meanwhile, increasing $K$ from $3$ to $5$ incurs negligible additional memory, with peak usage remaining at approximately $5.7$ GB on both datasets. This implies that when facing data for which convergence is slow, $K$ can be raised at negligible additional memory cost.}\rev{Nevertheless, $K$ can be increased when additional iterations are required, with negligible additional memory cost.}

\textbf{Impact of Input-Dependent Neuronal Parameters.}
Table~\ref{tab:input_conditioned_ablation} shows the contribution of input-dependent neuronal parameters. We remove the parameter generator and replace per-step $\tau_i$ and $R_i$ with learnable scalars shared across all steps. Accuracy drops from $91.25\%$ to $80.66\%$ on sequential CIFAR10 and from $66.59\%$ to $54.52\%$ on sequential CIFAR100. These results confirm that input-dependent parameter generation can significantly improve neuron performance.

\section{Conclusion}
We proposed FidLIF, a framework that enables parallel training while preserving high-fidelity hard-reset LIF dynamics and step-by-step inference.
FidLIF applies Newton's method to recover the membrane-potential trajectory iteratively in parallel. In the backward pass, FidLIF's reverse adjoint scan achieves BPTT-equivalent gradients without passing through the forward iterations,  making memory cost almost independent of the number of iterations. FidLIF also supports input-dependent neuronal parameters, improving the representational ability.
Experiments show that our Triton implementation achieves significant speedup and memory savings, particularly on long sequences. In real sequential benchmarks, FidLIF achieves competitive accuracy within 3 iterations, and the trained model supports step-by-step inference without extra conversion.
FidLIF thus combines high-fidelity hard-reset LIF dynamics with sequence parallelism, allowing SNNs to take advantage of modern accelerators during training while executing step-by-step during deployment. These features help SNNs strike a better balance between biological plausibility, efficiency, and edge deployment.

\bibliography{aaai2027}


\end{document}